# Modular Adaptive System Based on a Multi-Stage Neural Structure for Recognition of 2D Objects of Discontinuous Production

**Topalova, I.**
Department of Automation of Discontinuous Production
Technical University Sofia, Bulgaria
itopalova@tu-sofia.bg

*Abstract: This is a presentation of a new system for invariant recognition of 2D objects with overlapping classes, that can not be effectively recognized with the traditional methods. The translation, scale and partial rotation invariant contour object description is transformed in a DCT spectrum space. The obtained frequency spectrums are decomposed into frequency bands in order to feed different BPG neural nets (NNs). The NNs are structured in three stages - filtering and full rotation invariance; partial recognition; general classification. The designed multi-stage BPG Neural Structure shows very good accuracy and flexibility when tested with 2D objects used in the discontinuous production. The reached speed and the opportunuty for an easy restructuring and reprogramming of the system makes it suitable for application in different applied systems for real time work.*

## 1. Introduction

The translation, scale and rotation invariant recognition of 2D objects, which have overlapping classes in the feature space is a question of present interest in many scientific fields with applicability in: the visual robot systems; for the automated assembly mounting; in the visual checking systems etc.The usage of neural nets in this case have gained advantage over the traditional methods for invariant recognition, because of their capability to set the boundaries between the classes precisely.

There are two main approaches for invariant pattern classification using neural networks. The first approach uses invariant feature extraction and then employs neural networks to perform the classification on the feature values. Examples of such invariant descriptors are the extraction of image information using regular moments (Tsirikolias, K. & Mertzios, B.,1993), (Sluzek, A.,1995), Zernike moments (Khotanzad, A. & Hong, Y.,1990), Fourier-Mellin descriptor (Grace, A. & Spann, M.,1991), Fourier-wavelet descriptor (Guangyi, Ch. & Tien, D.B.,1999), Wavelet shape descriptors (Chang, G.C. & Jay Kuo, C.C.,1996), (Dinggang, Sh. & Horace, H.S.,1999), Wavelet transformation zero-crossing representation (Tieng, Q.M. & Boles, W.W.,1998). After the extraction of these invariant descriptors, their discrete components may feed neural networks to perform the recognition. The methods using regular moments are not efficient because they are highly noise-sensitive. Zernike moments outperform other kinds of moments, especially when noise is added but another general problem is that their calculation is very complicated and this slows down the preprocessing of the input vector. One of the main advantages of Fourier-based methods is that they can be applied efficiently using Fast Fourier Transform (FFT) techniques. For example Fourie-Mellin descriptors perform well under noise, but unlike moments are not translation-invariant. That is the reason the input pattern should be centralized before these descriptors are calculated . A negative point is that FFT requires complex-valued arithmetics. The disadvantage of Wavelet descriptor is that it isn't translation invariant and its calculation is a process of high computational complexity. The invariant contour descriptor (Kyoung, S.R. & Kweon, I.S.,1998). works on calculating determinants, which slows down the preprocessing stage in the recognition system. The second approach uses higher-order neural networks to produce invariance. The methods (Perantonis, S. & Lisboa R.,1992), (Spirkovska, L. & Reid, M.,1992) rely on the presence of symmetries among the network connections. The effect is, that it can learn the invariance. Weight sharing and higher-order neural networks outperform other invariant classifiers, but there is a severe problem with the amount of computational space they require, particularly in the case of third order networks.In our research we use the first approach, representing the object boundary and its inside closed contours by one-dimensional signals. We use a modified radial function, which gives a translation, partial rotation and scale invariant object representation after normalization. The obtained DCT (Discrete Cosine Transform) coefficient spectrums of the modified radial functions are decomposed into different frequency bands in accordance to a homogeneity standard. The different frequency bands that are achieved are input patterns for a three-stage BPNN (Backpropagation Neural Network) structure. The first stage BPG (Backpropagation) neural nets make the preprocessing (filtering) of the input patterns and give a full rotation invariant object description . The outputs of the first stage feed the next BPNNs which form the stage of partial recognition. The recognized object components on the second stage are composed in a vector that feeds the next BPNN. This BPNN forms the third stage - general classification. We use first-order BPNNs of feed forward type.

The main purpose of the research is the creation of invariant recognition of compact 2D objects with overlapping classes that can not be recognized precisely with the traditional methods. It is an extension of our experiments (Kountchev, R. & Topalova, I. ,1998) with precisely analysing the requirements for band separation, with another support vector for the filtering BPNNs, and with more experiments with different shapes. The invariant recognition is done at high recognition rate and at fast computation. In this research the proposed system is taught with six groups of objects. Each group has four classes. Each class is represented with exemplars, and noise is added to some of them . The recognition accuracy and the generalization of the system is tested with exemplars that don't belong to the training set. The experimental results are compared with the results obtained in other similar works. The comparison shows that the proposed system gives more stable and accurate results as well as shorter recognition time even for noisy images.

## 2. Description of the proposed system

The proposed recognition system is shown in Fig.1. The recognition system is designed to recognize 2D objects of m classes.The binary images of the training set and images that don't belong to the training set are stored in data base $DB_0$. The designed system consists of two parts. The first one is the preprocessing part (Fig.1a) which gives invariant object description. It includes block 1- extraction of translation and partial rotation invariant geometrical descriptors, block 2- scale normalization and DCT, block 3-amplitude normalization and division in different frequency bands. The obtained invariant descriptors are stored in the $DB_1$ and feed the BPNN structure, which forms the second part of the recognition system (Fig1b). It consists of three stages: filtering and full rotation invariance, partial recognition and general classification. The first and second stages are designed of separate BPNNs, which numbers are equal to the number of the vectors obtained in $DB_1$. The third stage consists of one BPNN only. The BPNN structure works in two modes which alternate through the multiplexers (Mux1 and Mux2). The first mode A is learning of the BPNN structure consecutively stage by stage. In this mode the $BPNN_i$ in each stage are trained with already recognized vectors at the previous stage, stored in the relevant DBs. The second mode B is recognition of exemplars that do not belong to the training set to prove the generalization of the system. The functional blocks will be described consecutively.

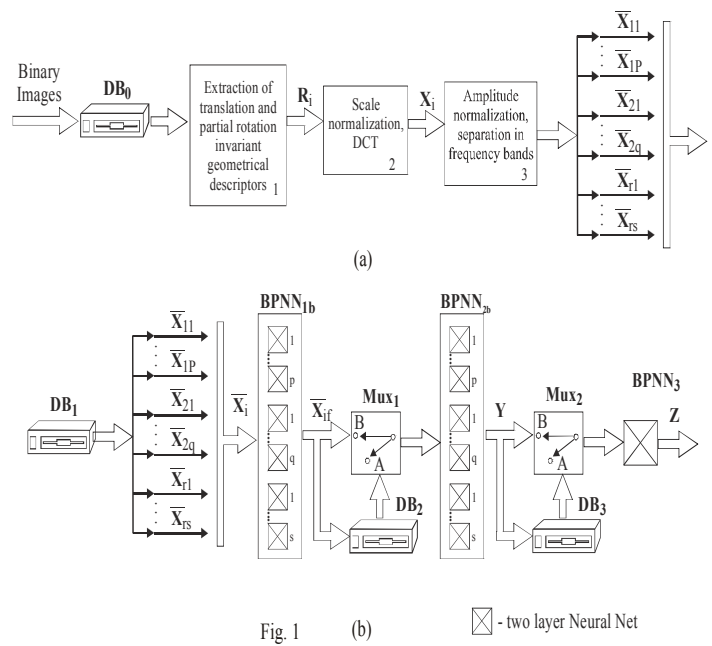

Fig.1. Block diagram of the proposed system

## 3. Description of the functional blocks

*3.1 Extraction of translation and partial rotation invariant geometrical descriptors*
The boundary of an object can be approximated by an ordered sequence of angularly equispaced vectors projected between an arbitrary reference point and the boundary points. This is often called the radial function (profile) (RP) - $R(\varphi)$ of an object (Kyoung, S.R. & Kweon, I.S.,1998). The RP reduces the dimensionality of the problem from a 2D object contour to 1D signal and has the partial rotation invariance property because it depends on the starting point on the object contour. If any of the radial vectors intersects the object boundary more than once, the function $R(\varphi)$ will be multivalued and can not be directly used to represent the contour. To overcome this restriction, we describe the contours of objects using the same vector, R($\ell$) but expressing them as a function of the parameter $\ell$, which varies from zero to the length of the perimeter of the object, L (Fig.1a, Bl.1). In the discrete case, the vector of the i-th contour,

$$R_i = \left[ r_{i1}, r_{i2}, ..., r_{ij}, ... r_{in} \right]^t$$

of the object centroid $x_c, y_c$ in relation to each contour point j, is calculated as

$$r_{ij} = \sqrt{(x_c - x_{ij})^2 + (y_c - y_{ij})^2}, \quad (1)$$
for $j = 1, 2, ..., n$

where n is the number of the relevant outline points. The RPs of the attached to the object inside closed contours are calculated as in (1), related to the same $x_c, y_c$ but with $x_j, y_j$ coordinates of the j-th hole's contour point. In this way RPs are translation invariant and partial invariant to rotation of the object in relation to the coordinate system of the picture. So the obtained vectors $R_i$ (i=1) about object boundary and it's attached r holes (I = 2,3,...,r+1.) are partial rotation and translation invariant.

*3.2. Scale normalization and DCT*
To obtain a scale invariant object description an interpolation and decimation of the contours have been made (Fig.1a, Bl. 2). Thus the length of the RPs for the object, and its attached holes is reduced to $N_i$ and is chosen in advance. The minimal value of $N_i$ is limited by the requirements of reconstructing the form of the object with a preliminary chosen accuracy. The maximal value of $N_i$ is limited in accordance with the requirements of the BPNNs regarding the computational time. A scaling coefficient p = $N_i$/n is used, then rounded and converted into a common fraction p = $N_i$/n = c/d. Then the RFs are increased 'd' times and decreased 'c' times to obtain $N_i$ components of the RFs. Then the vectors $R_i$ are transformed to the vectors $X_i$ after a DCT transformation, $X$=[C]$R_i$ where [C] is the matrix of DCT with a size of $N_i$. So the discrete frequency spectrums (FS)

$$X_i = \left[ x_{i1}, x_{i2}, ..., x_{iN_i} \right]^t$$

about object outline and its attached holes are obtained. Then the homonymous coefficients of all holes FSs attached to one and the same object, are compared. The first pair of coefficients for which a difference exists are markers for the respective hole and are arranged in rising order. Every hole in each object (class) is marked as 'first', 'second', 'third', etc. in accordance to the place of its marker in this range. So the hole's FSs are stored correctly in $DB_1$ independent on the object's rotation.

The dependance on the starting point of the object contour (for different rotations of an object) is reflected on the RF like a cyclic translation h of RF where $h = 0, 1, ..., N_i - 1$
The translation of RF works on the DCT coefficients as follows: $X_{ih} = [A_i] X_i$ Where $[A_i]$ has a size of $[N_i]$
and $a_i(u,h) = \left[ \cos(\beta) - \sin(\beta) \dfrac{s_i(u)}{x_i(u)} \right]$

$$h = 0, 1, ..., N_i - 1 \quad u = 0, 1, ..., N_i - 1 \quad (2)$$

$$s_i(u) = \sqrt{\dfrac{2}{N_i}} \sum_{j=0}^{N_1-1} r_i(j) \sin \dfrac{(2j+1)u\pi}{2N_i}$$
$$\beta = \dfrac{hu\pi}{N_i} \quad (3)$$
$$x_i(u) = \sqrt{\dfrac{2}{N_i}} \sum_{j=0}^{N_1-1} r_i(j) \cos \dfrac{(2j+1)u\pi}{2N_i}$$

where $x_{ih}$ is the DCT coefficient for the translated RF; $x_i(u)$ - for a base (not translated) RF; u is the current DCT frequency. The equations (2) and (3) show that $X_{ih}=X_i$ for h=0. As $x_i(u)$ and $s_i(u)$ are decreasing with increasing u and $s_i(u)$ is decreasing more rapidly then $x_i(u)$, the influence of $a_i(u,h)$ over $x_i(u)$ will be more significant about high frequencies.

*3.3. Amplitude normalization and separation in bands*
To reduce the amplitude of the $X_i$ components to the range of argument of the BPNN activation function, a modulus normalization of $X_i$ is applied (Fig.1a, Bl. 3). The coefficients of the amplitude normalized frequency spectrums for object boundary and for the attached holes are defined as follows:

$$\overline{X_i}(j) = \dfrac{X_i(j)}{\left| \sqrt{\sum_{j=1}^{N_i} x_i^2(j)} \right|} \quad for \ i = 1, 2, ..., r+1 \quad (4)$$

Then we separate the obtained normalized frequency spectrums in different bands to facilitate the work of the BPNNs. First the training time reduces when the number of input nodes is not too large. So the separation in bands reduces the lenght of the vectors, that feed the BPNNs. Second as the obtained DCT spectrums are decreasing, the separation aims to group these components of the vectors, that have small variance in a separate band. And third keeping the variance small in the range of each band leads to grouping the frequencies, for which $a_i(u,h)$ has a greater modifying effect in a separate band. So feeding each BPNN with a FS band having components with similar valued amplitudes will result in improvement of their adaptive capabilities. The frequency description $\mathbf{X}_i$ is separated into different frequency bands checking a homogeneity standard. So the obtained frequency bands:

$$\overline{\mathbf{X}_{ib}} = \left[ \overline{x}_{ib_1}, \overline{x}_{ib_2}, \ldots, \overline{x}_{ib_{Ni}} \right]^t$$

stored in $DB_1$ are as follows:

$\overline{X}_{1b}$ for object boundary, separated in p bands i.e. where $b = 1,...,p$ and i=1,

$\overline{X}_{2b}$ for the 'first' holes, separated in q bands i.e. where $b = 1,...,q$ and i=2,

$\overline{X}_{rb}$ for the "last" holes - separated in s bands i.e. where $b = 1,...,s$ and i=r+1.

*3.4. Multi-stage BPNN recognition structure*
The object description is decomposed in frequency bands which serve as input patterns ($DB_1$) to the BPNN recognition structure. This structure consists of three stages - filtering and full rotation invariance, partial recognition and general classification (Fig.1b). As each frequency band feeds a separate BPNN, their number in stage1 and 2 is equal to the number of the obtained in $DB_1$ vectors. Stage 3 consists of only one BPNN. Each stage works in two modes - A and B which alternate through the multiplexers (Mux1 and Mux2). Mode B is a recognition of exemplars that do not belong to the training set to prove the generalization of the system.
In A mode the BPNN structure is trained consecutively stage by stage. First $BPNN_{1b}$ are trained with a chosen set of vectors stored in $DB_1$. Then the same vectors are recognized (time interval $T_1$, Fig.2) through $BPNN_{1b}$ and are stored in $DB_2$ in order to serve as training set for the next $BPNN_{2b}$. After training the $BPNN_{2b}$, the $DB_2$ vectors are recognized (time interval $T_2$, Fig.2) through $BPNN_{2b}$ and are stored in $DB_3$ to serve as training set for the next $BPNN_3$ which is trained and then recognizes the vectors (time interval $T_3$, Fig.2). So in the training A mode, the different BPNN stages consecutively go through learning(L) and recognition(R) as it is shown in Fig.2.
The input vectors in $DB_1$ feed $BPNN_{1b}$. They are taught with 20 exemplars that have different rotation angles i.e. these vectors are modified with $a(u,h)$ regarding a vector for a base object, that isn't rotated. The base object is arbitrary chosen. The support vectors at the outputs of $BPNN_{1b}$ represent the average value of the corresponding vectors of different noisy base objects of each class. So these BPNNs are designed for filtering the noisy vectors and for obtaining at their outputs a full rotation invariant description. The filtered vectors $\overline{X}_{ibf}$ feed the next stage $BPNN_{2b}$. Each one of these BPNNs have a number of output nodes equal to the number of the taught classes - m. Their support vector has only one component with maximum value, which corresponds to the current teachable class. These BPNNs form the stage of partial recognition. The vector $\mathbf{Y}$ combines the components of the recognized vectors at the outputs of the $BPNN_{2b}$ as input vector for a general classification. The output vector $\mathbf{Z}$ has only one component with maximum value, which corresponds to the taught class. All BPNNs have identical structures and belong to both feedforward and two layer type. The hyperbolic tangent function is chosen as nonlinear activation function.

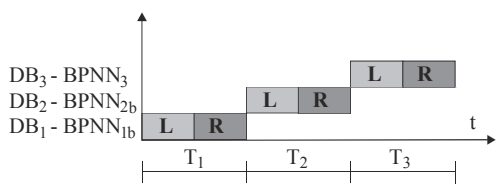

Fig.2. Training the BPNN structure consecutively stage by stage in mode A.

## 4. Experiments and simulation results

The proposed structure is tested with six groups of objects where each group has four classes. These groups with a reference representative of each class are shown in Fig.3.

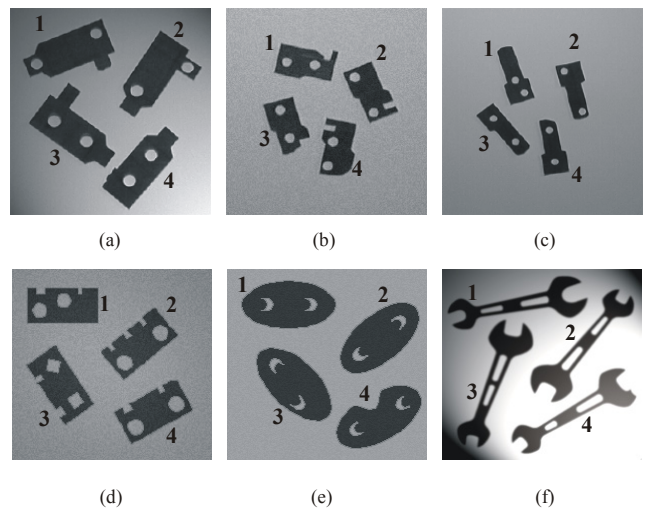

Fig. 3

Fig.3. Six groups of tested objects

All the representatives have 256 gray levels and a resolution of 256x256 pixels. They have the same global form in the group but with local alterations in each class. The training set includes 20 representatives of each class. They are obtained through modifying the reference object with variations in the rotation degree, in the position, in the size and for some representatives in the amount and kind of the added noise.

We obtain two frequency bands (low $X_{1I}$ and high $X_{1III}$) for the boundary and no need for division for the attached holes. The calculated normalized $X_1$ of class 2 (Fig.3f) for the base object -$X_{1g}$ and for the rotated representative of the same class - $X_{1h}$ is shown in Fig. 4a. $X_{1g}$ and the filtered FS $X_{1f}$ for the same object are shown in Fig. 4b. For better visualization we choose

$$B = sgn(\bar{X}_I)\sqrt{|\bar{X}_I|}$$

So it is visible that the influence of object's rotation over $X_1$ is ignored in $X_{1f}$. The variance of $a(u,h)$ for different rotation angles of the same object is represented in Fig.5. It is visible, that the variances of $a(u,h)$ become greater with increasing u. That means, the FS is sensitive to the object's rotation in its high frequency area and practically insensitive in its low frequency area. As FS of $X_1$ is decreasing for high frequencies, therefore the greater influence of $a(u,h)$ with the increase of u does no cause a greater variance of the FS's components in the obtained high frequency area.

That means, the chosen standard for homogeneity enables such a division in bands, that the influenced by $a(u,h)$ frequencies would be grouped in a different band. As each band feeds a different BPNN, the closer in value amplitudes of $X_1$ components in the band would facilitate the adaptive capabilities of the relevant BPNN, which decreases the learning time.

The calculated degree of overlapping $S_{kl}$ between the classes in Fig.3f are represented in Table1. in order to illustrate clearly the overlapping at the input of the BPNN structure. The calculated $S_{kl}$ between the classes in Fig.3f at the output of the BPNN structure are represented in Table 2. Tables 1 and 2 show that high overlapped classes at the input of the BPNN structure are well separated at the output of the structure.

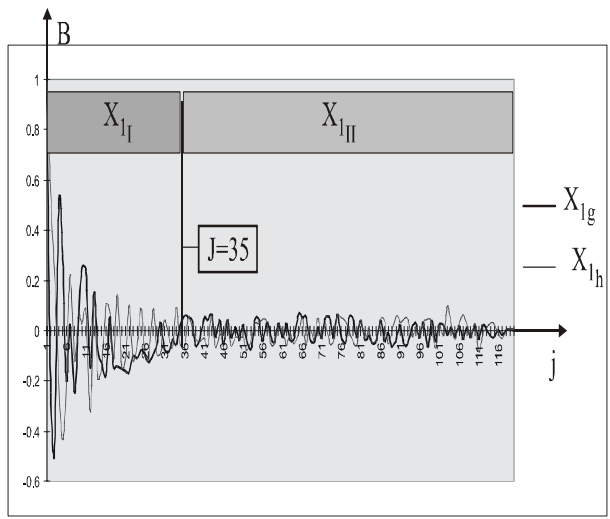

a)

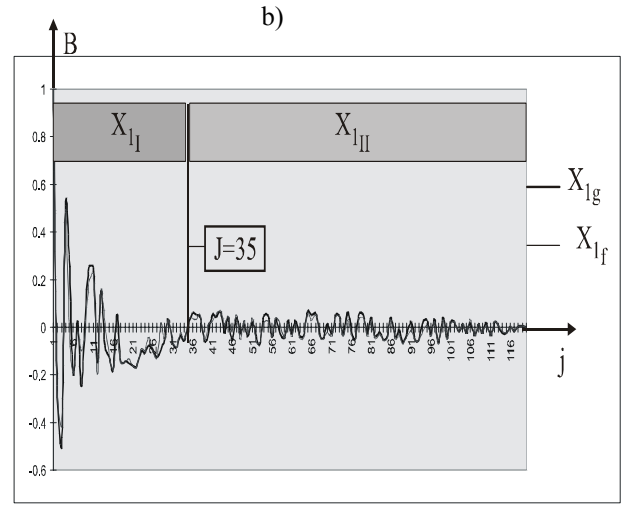

Fig.4. The calculated normalized FS of class 2; (a) about a base object and for a rotated representative of the same class ; (b) the filtered FS for the same object.

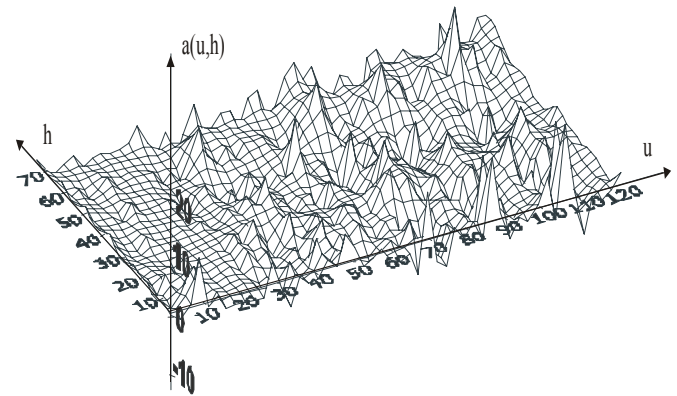

Fig.5. Variances of $a(u,h)$ depending of the DCT frequency u and of the translation h of RF

| $S_{kl}$ | l=2 | l=3 | l=4 | $S_{kl}$ | l=2 | l=3 | l=4 |
|---|---|---|---|---|---|---|---|
| k=1 | 0.56 | 0.5 | 0.39 | k=1 | 0.01 | 0.003 | 0.001 |
| k=2 |  | 0.53 | 0.46 | k=2 |  | 0.004 | 0.002 |
| k=3 |  |  | 0.41 | k=3 |  |  | 0.002 |

Table 1. The calculated $S_{kl}$ between classes k and $\ell$ at the input of the BPNN structure

Table 2. The calculated $S_{kl}$ between classes k and $\ell$ at he output of structure the BPNN

50 exemplars of each class were generated with different scale factors, different rotations and positions. To some of them 20, 25 and 30dB Gaussian noise or 5% blur was added. In mode B these objects were recognized in order to prove the generalization of the system.

A recognition rate of 76-82% is obtained at the outputs of the partial recognition stage (vector **Y**) and of 88-97% at the outputs of the general classification stage (vector **Z**) about all classes in the tested groups for 50 representatives of each class in each group. The recognition time including the preprocessing stage is at an average of 0.9 sec. for the different classes in the groups shown in Fig. 3.

## 5. Comparison of results

It is interesting to compare the results obtained by other researchers, when the objects represent overlapping classes. We investigated the recognition ability of the proposed representation against the use of Fourier descriptors and against the zero-crossing Wavelet representation of the contours given in (Tieng, Q.M. & Boles, W.W.,1998). The authors use minimum distance classifier. To compare the results we have generated the testing set (Fig.6) that is used in (Tieng, Q.M. & Boles, W.W.,1998).

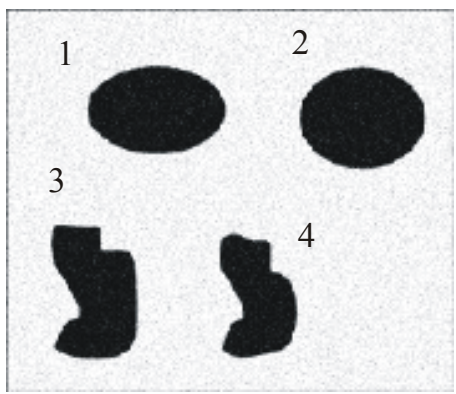

Fig.6. Training set in (Tieng, Q.M. & Boles, W.W.,1998)

The classes 1, 2 and 3, 4 in Fig.6 are overlapping with $S_{12}$=0.49 and $S_{34}$= 0.51. We have generated one hundred test runs of these objects with Gaussian noise of 20, 25 and 30 dB. To compare the results we trained the structure in Fig.1 with 20 exemplars of each class and then the 100 representatives were recognized. We use the comparative table represented by the authors (Tieng, Q.M. & Boles, W.W.,1998) with our results in addition. Table 3 lists the number of mismatches when Fourier descriptors (FD) are used, when different kinds of dissimilarity functions D1,...,D4 by zero-crossing Wavelet representation are used and when the proposed in this paper (RP, DCT) method with a three-stage BPNN recognition structure is applied. It is visible that the number of mismatches per 100 exemplars is zero in our case and if zero-crossing Wavelet representation is used but only if a false zero-crossing elimination algorithm is applied as proposed by the authors. But this algorithm is time consuming, which slows down the preprocessing. The authors of (Tieng, Q.M. & Boles, W.W.,1998) obtain 18.6 sec processing time for the construction of the representation and for the recognition procedure. In our case this time is at an average of 0.9 sec.

| No. miss. | FD | D1 | D2 | D3 (a) | D3(b) | D4(a) | D4(b) | RP, DCT |
|---|---|---|---|---|---|---|---|---|
| 30 dB | 29 | 10 | 6 | 7 | 1 | 43 | 2 | 0 |
| 25 dB | 15 | 5 | 2 | 23 | 1 | 28 | 0 | 0 |
| 20 dB | 15 | 3 | 2 | 23 | 0 | 21 | 0 | 0 |

Table 3. Number of mismatches in comparison to the results in (Dinggang, Sh. & Horace, H.S.(1999) and (Tieng, Q.M. & Boles, W.W.,1998)

The authors (Dinggang, Sh. & Horace, H.S.,1999) use wavelet moment invariants to recognize 2D objects with overlapping classes and compare the results with that of Zernike's moment invariants. In this case the authors use minimum distance classifier too. They use the two shapes shown in

Fig.7, which are rather similar. We have calculated their $S_{12}$ as $S_{12}$=0.431. The authors (Dinggang, Sh. & Horace,

H.S.,1999) obtain 100% recognition rate. In comparison the highest classification rate obtained by Zernike's moment invariants is 98%. In our case we compare the results obtained for objects in Fig.3f, because their shapes are similar to that used by (Dinggang, Sh. & Horace, H.S.,1999) - Fig.7. Comparing the degree of overlapping between the classes in Fig.3f (shown in Table 1) with that calculated for tool 1 and tool 2 it is evident that in our case the overlapping is greater. Considering the greater degree of overlapping and the testing of the system with noisy objects (till 30dB) in our case we may conclude that our system gives more accurate and stable results (Table 2,3).

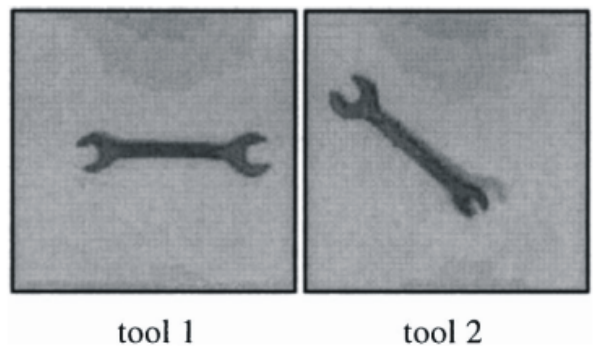

Fig.7. Training set in (Dinggang, Sh. & Horace, H.S.,1999)

The discriminability in these three cases is shown in Fig. 8a with Zernike's moment invariants, Fig. 8b with wavelet moment invariants, Fig. 8c with RP, DCT and BPNN in our case. For example the mean square deviation about class 2 in Fig. 8c is 2Fz =0.12 and , 2Fz=0.41  while for tool2 in Fig. 10b, 2Fx=6.6  and 2Fx=5.9.

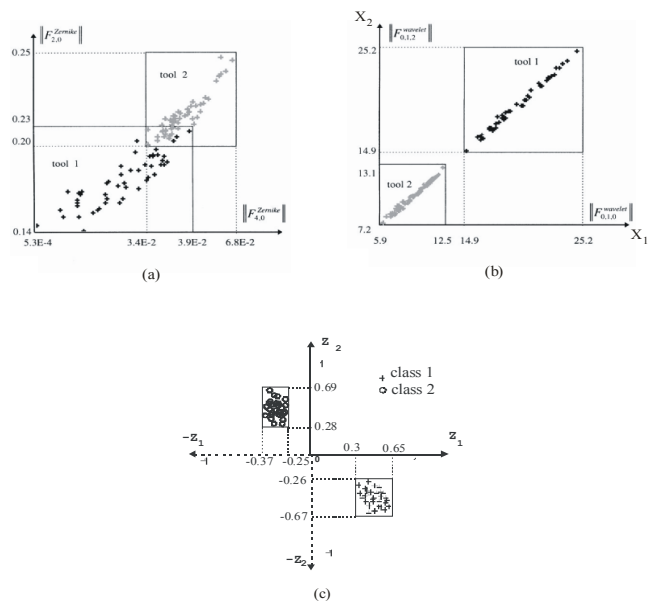

Fig.8. Degree of overlapping –
a- with Zernike's moment invariants (Dinggang, Sh. & Horace, H.S.(1999)
b – with wavelet moment invariants (Tieng, Q.M. & Boles, W.W.,1998)
c – RP, DCT – our case

## 6. Conclusions

The achieved accuracy and the opportunity for different combinations of the designed neural structures, depending of the kind and complexity of  the recognized objects (states), is a precondition for design of flexible recognition systems and their use to solve different applied tasks.
The separation of BPG structures in three functional groups and teaching them stage by stage, using already recognized components of the input vectors, leads to increasing the recognition accuracy.
The use of a different-band coding of the input vectors reduces the total number of input elements of each BPNN and facilitates the adapting capabilities of neural nets. The identical structures of the BPNNs give an opportunity to use the same running weight matrixes and to perform parallel computations. Thus the needed memory resources are reduced as well as the computational time.
The proposed system provides a possibility for a high recognition rate of 100% for the tested objects  The method is universal for compact 2D objects with different form and is distinguished as considerably noise resistant.
The results of this study outperform that of the methods given in (Dinggang, Sh. & Horace, H.S.(1999) and (Tieng, Q.M. & Boles, W.W.,1998) regarding the recognition accuracy as well as the noise resistance.
The presented system does not require any preliminary knowledge, because it is a self-teaching system. Due to the ability to adjust the method for different kinds of 2D objects it can be applied in different fields of learning, in computer vision  and real time working systems (Neschkov, T. & Topalova, I.(2001). .